\documentclass[10pt,twocolumn,letterpaper]{article}

\usepackage[pagenumbers]{iccv} 

\usepackage{utfsym}
\usepackage{subcaption}
\usepackage{graphicx}
\usepackage{multicol}
\usepackage{multirow}
\usepackage{arydshln}
\usepackage{wrapfig}
\usepackage[subtle,mathdisplays=normal, wordspacing=normal]{savetrees}
\usepackage{wrapfig,lipsum,booktabs} 
\usepackage{pifont}  
\definecolor{iccvblue}{rgb}{0.21,0.49,0.74}
\usepackage[pagebackref,breaklinks,colorlinks,allcolors=iccvblue]{hyperref}

\def\paperID{6824} 
\def\confName{ICCV}
\def\confYear{2025}

\title{Beyond Spatial Frequency: Pixel-wise Temporal Frequency-based\\ Deepfake Video Detection}

\author{
Taehoon Kim$^{1}$,\hspace{1cm}Jongwook Choi$^{1}$,\hspace{1cm} Yonghyun Jeong$^{3}$,\hspace{1cm} Haeun Noh$^{1}$,\\
Jaejun Yoo$^{4}$,\hspace{1cm} Seungryul Baek$^{4}$,\hspace{1cm} Jongwon Choi$^{1,2}$\thanks{Corresponding author} \\
{\small $^1$GS. of AI, Chung-Ang Univ, Korea}\hspace{1cm}
{\small $^2$Dept. of Advanced Imaging, Chung-Ang Univ, Korea}\\
{\small $^3$NAVER Cloud, Korea}\hspace{1cm}
{\small $^4$AI Graduate School, UNIST, Korea}\\
    {\tt\small
    \{kimth, cjw\}@vilab.cau.ac.kr,
    yonghyun.jeong@navercorp.com,
    nhe8354@vilab.cau.ac.kr,}
    \\
    {\tt\small\{jaejun.yoo, srbaek\}@unist.ac.kr,
    choijw@cau.ac.kr}
}

\begin{document}
\maketitle
\AddToShipoutPicture*{%
     \AtTextUpperLeft{%
         \put(0,30){
           \begin{minipage}{\textwidth}
              \footnotesize
                Preprint version; final version will be available at \url{https://openaccess.thecvf.com}\\
                The International Conference on Computer Vision (ICCV) (2025)\\
           \end{minipage}}%
     }%
}
\begin{abstract}
We introduce a deepfake video detection approach that exploits pixel-wise temporal inconsistencies, which traditional spatial frequency-based detectors often overlook. Traditional detectors represent temporal information merely by stacking spatial frequency spectra across frames, resulting in the failure to detect temporal artifacts in the pixel plane. Our approach performs a 1D Fourier transform on the time axis for each pixel, extracting features highly sensitive to temporal inconsistencies, especially in areas prone to unnatural movements. To precisely locate regions containing the temporal artifacts, we introduce an attention proposal module trained in an end-to-end manner. Additionally, our joint transformer module effectively integrates pixel-wise temporal frequency features with spatio-temporal context features, expanding the range of detectable forgery artifacts. Our framework represents a significant advancement in deepfake video detection, providing robust performance across diverse and challenging detection scenarios.
\end{abstract}
\section{Introduction}
\label{sec:intro}
Deepfake video detection is getting increasingly in demand as facial synthesis techniques become more realistic and accessible.
Facial synthesis technologies have been used in various fields, such as movies, television shows, and advertisements, but have also led to a significant increase in social issues arising from fake news, unauthorized content, and explicit content~\cite{faceforensics,masood2023deepfakes}.

Recently, many deepfake video detection studies have adopted frequency-based detectors, focusing on the spatial aspects of images through Fourier transforms and filtering~\cite{SFDG,frepgan,LRL,NIR}. Early detectors identified artifacts in raw videos~\cite{faceforensics,facexray,cnngru,ftcn}, such as blurred boundaries, color inconsistencies, resolution differences, and flickering. In contrast, frequency-based detectors transform pixel values into frequency-aware features, making it easier to identify subtle artifacts for more effective detection. While pixel-level artifacts decrease with advanced synthesis methods, frequency-level artifacts robustly appear in generated videos due to their invisible characteristics, even in high-quality outputs.

\begin{figure}[t!]
    \centering
    \includegraphics[width=\columnwidth]{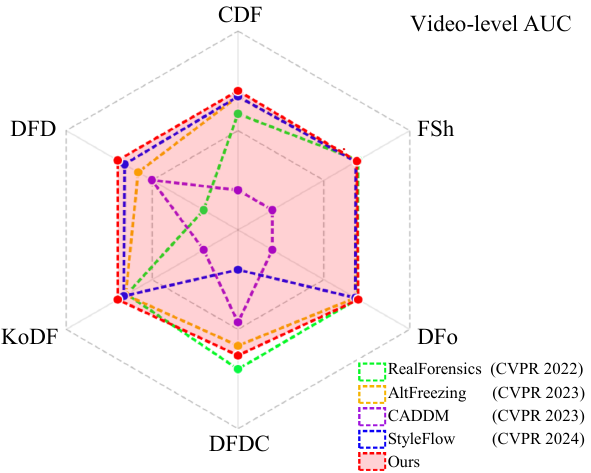}
    \vspace{-5mm}
    \caption{\textbf{Comparison with state-of-the-art methods.} Our approach leverages pixel-wise temporal frequency, which was not utilized in previous work. Ours outperforms the state-of-the-art methods across various unseen datasets, such as CDF~\cite{CDF}. For comparison, we trained only on FF++~\cite{faceforensics} and evaluated video-level AUC for unseen datasets.}
    \label{fig:teaser}
    \vspace{-7mm}
\end{figure}

While frequency-based deepfake video detectors can reveal invisible artifacts from synthesis models, the existing frequency-based detectors often overlook temporal artifacts. Video synthesis models frequently struggle with temporal inconsistency and unnatural frames. However, temporally-stacked spatial frequency spectra in existing detectors are insufficient for detecting temporal artifacts as they only consider changes in spatial artifacts and ignore pixel-wise temporal variations. 
Additionally, spatial frequency spectra relying on spatial integration are improper for detecting the temporal artifacts concentrated in specific regions.

To effectively utilize the temporal artifacts, we introduce a novel method for deepfake video detection that utilizes pixel-wise temporal frequency spectra. In contrast to the previous approach of simply stacking 2D frame-wise spatial frequency spectra for temporal association, we extract pixel-wise temporal frequency by performing a 1D Fourier transform on the time axis per pixel effectively identifying the temporal artifacts. Observing that temporal artifacts are often discovered in particular spots, we also developed an Attention Proposal Module (APM) trained in a weak-supervised scheme to extract regions of interest for detecting temporal artifacts. 
Our pixel-wise temporal frequency spectra effectively capture subtle artifacts present in deepfake videos, and these spectra are subsequently integrated with spatio-temporal context features through a joint transformer module.

By conducting experiments in various challenging deepfake video detection scenarios, including cross-domain, cross-synthesis, and robustness-to-perturbation experiments, our method demonstrates outstanding generalizability, as shown in Fig.\ref{fig:teaser}.
We additionally provide a comprehensive analysis of the role of temporal artifacts in deepfake video detection, further verifying the effectiveness of our approach.
Our contributions are summarized as follows:

\begin{itemize}
    \item {We introduce a novel approach to leveraging pixel-wise temporal frequency for deepfake video detection.}
    \item We designed an Attention Proposal Module (APM) to identify regions of interest for effectively detecting temporal artifacts.
    \item We present a joint transformer module that leverages temporal-frequency information for effective temporal artifact detection.
    \item Our method achieves state-of-the-art performance, demonstrating its effectiveness and generalizability.
\end{itemize}

\begin{figure}
    \centering
    \subcaptionbox{Temporal artifacts in Videos \label{fig:artifact}}[\columnwidth]{
    \centering
    \includegraphics[width=\columnwidth]{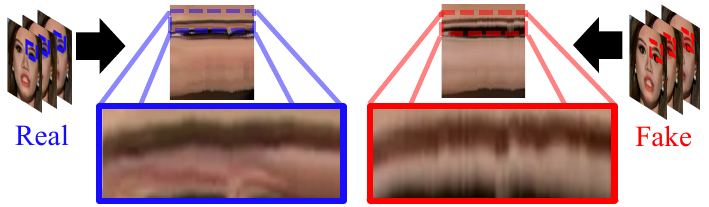}
    }
    \subcaptionbox{Frequency extraction method \label{fig:precompare}}[\columnwidth]{
 \hfill
    \centering
    \includegraphics[width=\columnwidth]{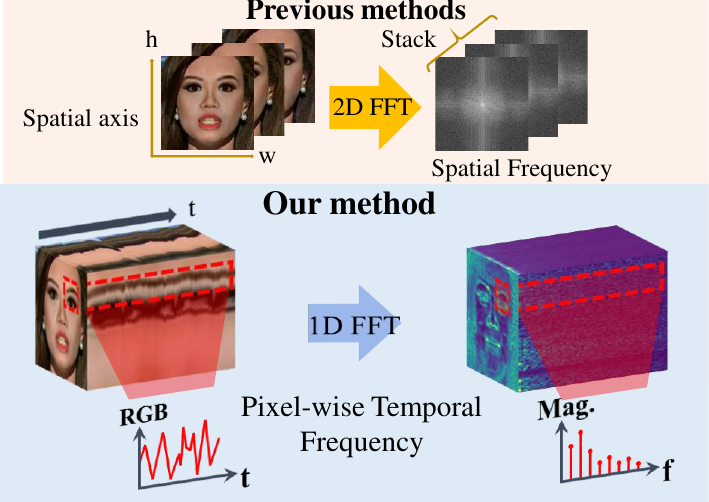}
    }
    \vspace{-3mm}
    \caption{\textbf{Visualization of temporal inconsistency and a mechanism of our pixel-wise temporal frequency.}
    In (a), we depict the motion of vertical slices along the time axis in the real and fake video, as suggested by the visualization in~\cite{NIR}. (b) presents the difference in frequency spectrum extraction between previous frequency-based methods and our pixel-wise temporal frequency method.}
    \label{fig:fig2}
    \vspace{-10mm}
\end{figure}
\section{Related Work}
\label{sec:related}
\noindent\textbf{Spatial artifact-based deepfake detection.}
Early research~\cite{Hog, head_pose, eyeblinking} attempted to detect awkward elements based on handcrafted features. However, with the advancement of deep learning, it has become possible to extract effective features for deepfake detection using deep learning models~\cite{Xception, Mesonet, Dfwild, mat, caddm}. 
Additionally, previous studies~\cite{facexray, sladd, sbi} have addressed artifacts from the post-processing of face swaps. 
Nevertheless, image-based methods face increasing challenges as blending techniques and generative models continue to improve.

\noindent\textbf{Temporal artifact-based deepfake detection.}
Early studies~\cite{cnngru, lrnet} exploiting temporal inconsistencies often employed recurrent models like GRUs~\cite{cnngru, GRU, lrnet} to capture temporal dependencies in deepfake videos. Deepfake detection methods~\cite{lips, realforensics} that use pre-trained networks trained on general motion information have demonstrated fairly good generalization performance. Recent studies~\cite{ftcn, altfreezing, RobustForensics,StyleFlow} have shown that reduced reliance on spatial information improves the generalization performance of deepfake detection models. 
While previous research has primarily focused on leveraging temporal artifacts at the RGB level, we propose a novel methodology that utilizes temporal frequency-level artifacts, which are imperceptible in conventional RGB-based analyses.

\noindent\textbf{Frequency-based deepfake detection.}
After Wang~\cite{wang2020cnn} found that GAN-generated images have easily recognizable peculiar frequency artifact patterns, many studies have tried to discover forgery patterns in frequency domains, improving performance. Some methods used various methods such as DCT or DWT to get frequency information from images~\cite{thinkfreq, f2trans, frepgan, bihpf, GanFingerprint,F3net}. HFF~\cite{highfreq} was also explored to distill high-frequency artifacts that discern deepfake images and concatenate the 2D frequency spectrum to distinguish manipulated faces from real ones. PEL~\cite{PEL} also integrated fine-grained frequency features and RGB images. SFDG~\cite{SFDG} suggested a spatial-frequency dynamic graph to learn the relation between spatial and frequency domains. Two-branch~\cite{twobranchrnn} and following studies~\cite{NIR, MGL, phasevideo} employed a strategy of stacking frame-wise spatial frequencies.
However, these methods predominantly analyze spatial frequencies within individual frames, which limits their ability to capture temporal information. To address this limitation, we propose a novel approach that explicitly targets temporal frequency artifacts by applying the Fourier transform along the temporal axis.

\section{Preliminary Analysis \label{sec:preliminary}}

In our observation, as shown in Fig.~\ref{fig:artifact}, we observed temporal inconsistencies, particularly around the eye area in fake videos, where a noticeable fine fluctuation is present in contrast to real videos.
Moreover, recognizing that such fine fluctuations predominantly occur in specific areas, we can find that the temporal artifacts can be more effectively detected by focusing on limited regions suspected to contain the inconsistencies.

To confirm the importance of temporal artifacts at specific regions in deepfake video detection, we conducted preliminary cross-synthesis experiments on four face forgery methods provided by FaceForensics++ (FF++)~\cite{faceforensics}.
To show the effectiveness of the pixel-wise temporal frequency spectrum, we designed a simple classification model to compare with existing spatial frequency-based detectors.
The classification model is designed with the basic ResNet-18 classifier that uses the pixel-wise temporal frequency spectrum as input. The method to extract the pixel-wise temporal frequency is represented in Subsection~\ref{subsec:frequency_extract} and visualized in Fig.~\ref{fig:precompare}.

As shown in Table~\ref{tab:landmark}, we can see that our simple model outperforms the image-based detectors such as Xception~\cite{faceforensics} and CNN-aug~\cite{wang2020cnn}, and spatial frequency-based detector HFF~\cite{highfreq}.
We can find that CNN-GRU~\cite{cnngru} using temporal information performs better than the spatial frequency-based and image-based detectors, but our simple model (\textit{whole-face}) even outperforms the CNN-GRU.

\begin{table}
    \centering\resizebox{\columnwidth}{!}{
    \begin{tabular}{l|ccccc}
\hline
\multirow{2}{*}{Method} & \multicolumn{5}{l}{Train on remaining three}          \\ 
                        & DF   & FS   & F2F  & \multicolumn{1}{c|}{NT}   & Avg  \\ \hline
Xception~\cite{faceforensics}                & 93.9 & 51.2 & 86.8 & \multicolumn{1}{l|}{79.7} & 77.9 \\
CNN-aug~\cite{wang2020cnn}                 & 87.5 & 56.3 & 80.1 & \multicolumn{1}{l|}{67.8} & 72.9 \\
CNN-GRU~\cite{cnngru}                 & 97.6 & 47.6 & 85.8 & \multicolumn{1}{l|}{86.6} & 79.4 \\ 
HFF*~\cite{highfreq}                    & 95.0 & 40.6 & 77.1 & \multicolumn{1}{l|}{71.5} & 71.1  \\ \hline\hline
whole-face                   & 96.5 & \underline{77.7}& \underline{93.9}& \multicolumn{1}{l|}{93.3} & 90.3 \\ \hdashline
right-eye               & 96.6 & 70.2 & 89.9 & \multicolumn{1}{l|}{92.8} & 87.5 \\
left-eye                & \underline{99.1}& \textbf{85.8} & 93.4 & \multicolumn{1}{l|}{94.3} & \textbf{93.2} \\
nose                    & \textbf{99.8} & 78.8 & 93.3 & \multicolumn{1}{l|}{95.1} & \underline{91.7}\\
lip                     & 84.5 & 59.4 &\textbf{ 94.3} & \multicolumn{1}{l|}{\textbf{96.4}} & 91.1 \\
right-side              & 93.9 & 62.4 & 86.8 & \multicolumn{1}{l|}{92.4} & 83.9 \\
left-side               & 96.1 & 70.7 & 93.3 & \multicolumn{1}{l|}{\underline{96.3}} & 89.1 \\ \hline
\end{tabular}%
}
\vspace{-2mm}
    \caption{\textbf{Preliminary experiments of cross-synthesis.} Above the double line is the result of the current method, while below is the result of training only with ResNet-18 using pixel-wise temporal frequency. The highest scores are highlighted in \textbf{bold}, while secondary results are \underline{underlined}. An asterisk (*) indicates our reproduced results.}
  \label{tab:landmark}\vspace{-5mm}
\end{table}

\begin{figure*}[t]
  \centering
  \includegraphics[width=1\textwidth]{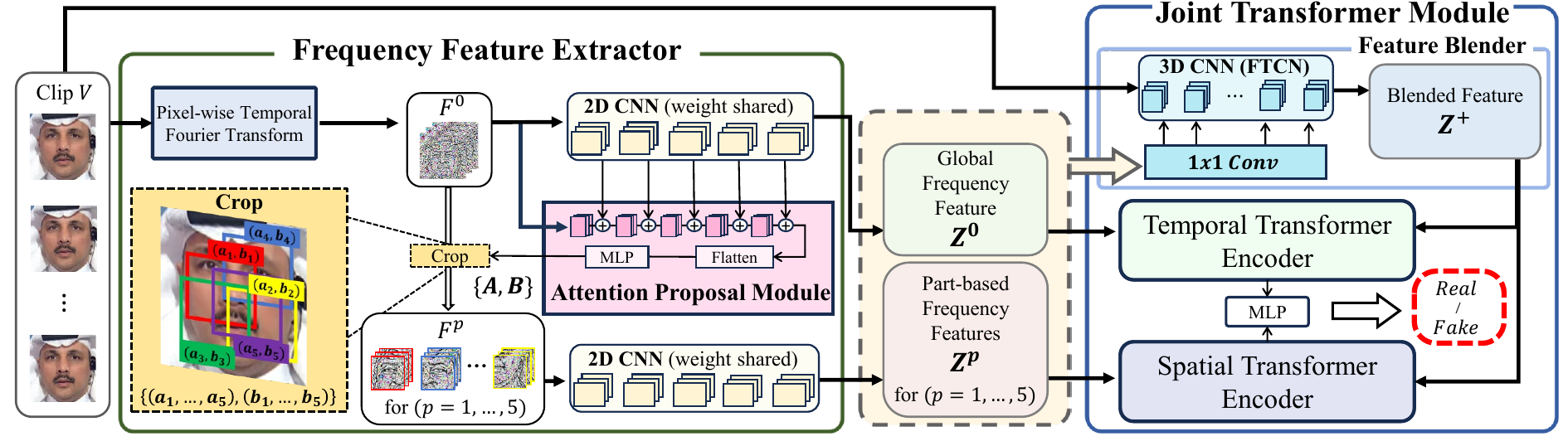}\vspace{-3mm}
  \caption{\textbf{The pipeline of our method.} For extracting temporal frequency $F^{0}$, the video clip $V$ is decomposed into temporal frequency components using the Fourier Transform. The frequency feature extractor obtains a part-based frequency feature $Z^{p}$ and a global frequency feature $Z^{0}$ using 2D ResNet and an attention proposal module. The part-based and global frequency features enter the feature blender to get a blended feature $Z^{+}$, and put the blended features, part-based frequency features, and global frequency features into a joint transformer module to classify real and fake.}\vspace{-5mm}
  \label{fig:framework}
\end{figure*}

Next, we conduct experiments to validate the importance of identifying specific regions where temporal artifacts appear.
The part-wise performance comparison under the double line in Table~\ref{tab:landmark}, \textit{whole-face} is a classification model using the entire facial region for the pixel-wise temporal frequency spectrum, and \textit{right-eye}, \textit{left-eye}, \textit{nose}, \textit{lip}, \textit{right-side}, and \textit{left-side} use the $66\times 66$ partially-extracted patches near the corresponding regions. 

From the results, we observe that the deepfake detector using a pixel-wise temporal frequency spectrum needs to concentrate on specific regions, and the regions vary according to the synthesis methods.
First of all, in every synthesis type, the part-based detectors present better performance than the detectors using entire facial images.
In addition, from the results where the different parts work best for the respective synthesis methods such as DF and NT, it has become evident that every generation algorithm contributes certain artifacts to their particular regions.

\section{Proposed Method}\label{sec:method}

In this section, we introduce the details of our proposed architecture, which consists of a frequency feature extractor and a joint transformer module.
The frequency feature extractor extracts temporal frequency features and determines the interesting regions to extract the features.
Then, the following joint transformer module first integrates the temporal frequency features extracted from the multiple interesting regions.
Finally, the transformer-based integration of the joint transformer module combines the spatial and temporal features for the final decision.
The overall framework is shown in Fig.~\ref{fig:framework}.

\subsection{Frequency Feature Extractor}
The frequency feature extractor extracts pixel-wise temporal frequency along the time axis from facial videos, feeding the frequency into 2D ResNet~\cite{resnet} to obtain global frequency features $Z^{0}$, where ``global" means whole facial region.
Then, the Attention Proposal Module (APM) extracts patches, which are fed back into the 2D ResNet used for $Z^0$ to obtain part-based frequency features $Z^p$ where $p \in \{1,2,3,4,5\}$.

\subsubsection{Global Frequency Feature Extraction.}
\label{subsec:frequency_extract}
We first divide the facial video into clips $V\in R^{C\times T\times H\times W}$ where consecutive $T$ frames are selected.
For each frame image $I^t\in R^{C\times H\times W}$ in $V$, we preprocess to obtain frame $\hat{I}^t$ by removing the dominant components using a median filter as:
\begin{equation}
    \hat{I}^t = \text{gray}(I^t - \text{Median}(I^t)),
    \label{equ:median_filter}
\end{equation}
where $\text{Median}(I)$ is an image filtered by a median filter and $\text{gray}(I)$ means the gray-scaling function from the colored image $I$ and we set T as 32.
Then, the pre-processed video clip is defined as $\hat{V}\equiv\{\hat{I}^1, \hat{I}^2, ..., \hat{I}^T\}$.

From $\hat{V}$, we extract the pixel-wise temporal frequency spectrum.
When we notate the temporal vector at the position of $(x,y)$ by $\hat{V}_{x,y}\in R^{1\times T}$, we can represent the pixel-wise temporal frequency spectrum at $(x,y)$ as follows:
\begin{equation}
    F_{x,y} = \mathcal{F}(\hat{V}_{x,y}),
    \label{equ:fourier_transform}
\end{equation}
where $\mathcal{F}(v)$ is a 1-dimensional frequency magnitude spectrum of input $v$.
Since the input vector is real-valued, we ignore the symmetric parts of the magnitude spectrum, so the shape of $F_{x,y}$ becomes $R^{1 \times T/2}$.
Then, we can integrate $F_{x,y}$ for every pixel to obtain the global frequency spectrum $F^{0}\in R^{1 \times T/2 \times H\times W}$.
Finally, $Z^{0}\in R^{D_H\times D_W \times D_C}$ is obtained by feeding $F^{0}$ into 2D ResNet.
For the first convolution of 2D CNN, the input channel is increased to $T/2$ to get the frequency feature.

\subsubsection{Part-based Frequency Feature Extraction.}
Inspired by \cite{racnn, macnn}, we design an Attentional Proposal Module (APM) that can effectively extract regions using features $Z^{0}$ from the initial convolution layer and every block of a 2D ResNet in a learnable manner.
APM takes the temporal frequency magnitude $F^{0}$ and the features $z^{0}_{(i)}$ extracted from $i$-th block of the 2D ResNet as input, regressing the coordinate set $(A,B)$ for the five parts to be focused on.
\begin{equation}
    [A,B] = \text{APM}\Big(F^{0}, z^{0}_{(0)}, z^{0}_{(1)}, ..., z^{0}_{(L)}\Big),
    \label{equ:apn}
\end{equation}
where $z^{0}_{(l)}$ is the feature from the $l$-th convolution layer of 2D ResNet, $L$ is the number of blocks in 2D ResNet, and $A = \{a_1, a_2, a_3, a_4, a_5\}, B = \{b_1, b_2, b_3, b_4, b_5\}$ that are the $x$ and $y$ center coordinates of the five parts presented by the APM, respectively.

APM generates a mask map $M_p$ for the $p$-th part using $a_p$ and $b_p$ to crop the local patch $F^{p}$ from $F^{0}$ through element-wise multiplication $\odot$ as follows:
\begin{equation}
    F^{p} = F^{0}\odot M_p(a_p,b_p),
    \label{equ:maskmul}
\end{equation}
where $M_p(a_p, b_p)$ is the $1
\times T/2\times H\times W$-dimensional array having $1$ for rectangular regions that have $(a_p-\theta,b_p-\theta)$ as the left-top position and $(a_p+\theta,b_p+\theta)$ as the right-bottom position and $0$ for remaining regions. 

The $\theta$ is the initial rectangular size that is empirically set to $44$. To obtain $M_p(a_p, b_p)$ in a differentiable manner from $a_p$ and $b_p$ which were estimated from the APM, we applied the below operation: 
\begin{equation}
    \begin{aligned}
        M_p(a_p, b_p) &= \Big[h(\mathbf{y} \text{-} (a_p\text{-}\theta)) \text{-} h(\mathbf{y} \text{-} (a_p + \theta))\Big]\\
                & \quad\times \Big[h(\mathbf{x} \text{-} (b_p\text{-}\theta)) \text{-} h(\mathbf{x} \text{-} (b_p + \theta))\Big],
    \label{equ:mask}
    \end{aligned}
\end{equation}
where $h(\mathbf{v})$ is a element-wise logistic function with the scale factor of $10$, $\textbf{x}$ is the $W$-dimensional vector having values from 0 to $W$, and $\textbf{y}$ is the $H$-dimensional vector having values from 0 to $H$.
For each $p$-th part features obtained, $F^{p} \in R^{T/2\times (2\times \theta)\times (2\times \theta)}$ is fed back into the 2D ResNet to extract the part-based frequency feature $Z^{p}$ for $p\in\{1,...,5\}$.

\subsection{Joint Transformer Module}

\subsubsection{Feature Blender.}
The feature blender integrates the raw features, obtained by feeding the raw clip $V$ into 3D ResNet~\cite{3dresnet,slowfast}, with the global and part-based frequency features.
We define the features extracted from $i$-th block of 2D ResNet fed by the $p$-th part-based frequency $F^{p}$ by $z^{p}_{(i)}$.

First, the various frequency features are integrated through $1\times 1$ convolution layers as:
\begin{equation}
    \tilde{z}_{(i)} = Conv_{1\times1}^f\Big(\sum^5_{p=1}{Conv_{1\times1}^p(z^{p}_{(i)})}+Conv_{1\times1}^{0}(z^{0}_{(i)})\Big),
    \label{equ:integration}
\end{equation} 
where $Conv_{1\times1}^0$, $Conv_{1\times1}^p$, and $Conv_{1\times1}^f$ are the separated output function of $1\times 1$ convolution layers.
$Conv_{1\times1}^f$ consists of two consecutive convolution layers, with the number of channels being halved at the first layer and then restored to the original amount at the second one, with a ReLU activation function placed between them. The weights of the second layer in $Conv_{1\times1}^f$ are initialized to zero. $Conv_{1\times1}^0$ and each $Conv_{1\times1}^p$ consist one layer.
By applying spatial interpolation to match the spatial dimension of $z^{p}$ with that of $z^{0}$, they can be summed.
The shape of $\tilde{z}_{(i)}$ is $R^{C_i\times H_i \times W_i}$ where $C_i$, $H_i$, and $W_i$ represent the size of channel, height, and width of $z^{0}_{(i)}$ ($i=0,...,3$).

Then, $\tilde{z}_{(i)}$ is added with the blended feature $Z^{+}_{(i)}$ from the $i$-th layer of the 3D ResNet $\Phi_i$ and passed into the $(i+1)$-th layer of the 3D ResNet $\Phi_{i+1}$.
\begin{equation}
    Z^{+}_{{(i+1)}} =  \Phi_{i+1}\big(\tilde{z}_{(i)} + Z^{+}_{(i)}\big),
    \label{equ:blender}
\end{equation} 
where $Z^{+}_{(0)}$ is the feature from the first convolution layer of 3D ResNet fed by raw video clip $V$.
Since the dimensions of $\tilde{z}_{(i)}$ and $Z^{+}_{(i)}$ are the same, they can be summed via broadcasting.
After the feature blender, we only utilize $Z^{+}_{(4)}\in R^{1024\times 16\times 14\times 14}$ as the integrated feature.

\subsubsection{Transformer-based Integration.}

To effectively integrate pixel-wise temporal frequency spectra with spatio-temporal context features, the joint transformer module processes three previously extracted features by: $z^0_{(4)}$, $z^p_{(4)}$(for $p = 1, ..., 5$), and $Z^{+}_{(4)}$.

We design the Spatial Transformer Encoder (STE) to integrate part-based frequency features with spatial artifact representations, enabling the model to effectively capture complex spatial relationships.
Similarly, the Temporal Transformer Encoder (TTE) leverages transformers to combine temporal frequency features with temporal context information, facilitating a more comprehensive understanding of temporal artifacts. 
Unlike \cite{ftcn}, which primarily considers long-term temporal consistency, our method explicitly incorporates pixel-wise temporal frequency representations, allowing for a more fine-grained detection of temporal artifacts.

$Z^{+}_{(4)}$ and $z^p_{(4)}$(for $p = 1, ..., 5$) are fed into STE for patial analysis, while $Z^{+}_{(4)}$ and $z^{0}_{(4)}$ are fed into TTE for temporal analysis.
Both STE and TTE consist of a single layer of the standard transformer encoder. 
To match the dimensions of $z^{0}_{(4)}$ with $Z^{+}_{(4)}$, we apply a linear projection.
The outputs from STE and TTE are then fed into the final classifier to obtain the final prediction $\hat{y}$.
Further detailed architecture is given in the supplementary material.

\subsection{Training Phase}
We employ the auxiliary classifier to learn the frequency feature extractor and STE.
We utilize two auxiliary classifiers $\phi^{g}$ and $\phi^{p}$ which are respectively fed by the global and part-based frequency features for binary classification of real and fake videos. 
We also use the auxiliary classifier $\phi^{sp}$, which is fed by the STE embedding features.
We then define the auxiliary classifier loss as:
\begin{equation}
\begin{aligned}
 \mathcal{L}_{\text{auxiliary}} &= \mathcal{C}_y\Big(\phi^{p}(Z^{P}_{(4)})\Big) + \mathcal{C}_y\Big(\phi^{g}(z^{0}_{(4)})\Big)\\ 
                                &\quad + \mathcal{C}_y\Big(\phi^{sp}\big(STE(Z^{+}_{(4)}, Z^{P}_{(4)})\big)\Big),
    \label{equ:phase1}
    \end{aligned}
\end{equation}
where $\mathcal{C}y$ is a binary cross-entropy loss function aligned with the corresponding ground-truth label $y$ and $Z^{P}$ is the concatenated features of $z^{p}_{(4)}$ for all $p$ (where $p = 1, ..., 5$).
The entire network is trained by using both the auxiliary and final classifier, with the training loss formulated as :
\begin{equation}
\mathcal{L}_{\text{final}} = \lambda\mathcal{C}_y(\hat{y}) + \mathcal{L}_{\text{auxiliary}},
\label{equ:phase2}
\end{equation}
where $\lambda$ is a weighting factor to focus on the final classifier loss $\mathcal{C}_y(\hat{y})$.
Thus, we have no specific training loss to designate the regions for the APM due to the lack of localization labels, but the final classification loss effectively drives the APM in an end-to-end manner to detect regions of interest for temporal artifacts.

\begin{table*}[t]

  \centering\resizebox{\textwidth}{!}{
  \begin{tabular}{l|c|c|c|ccccc|c}
\hline
Method                         &  Extra Dataset&Input Type  & Feature          & CDF            & DFDC  & FSh            & DFo  & DFD            & Avg            \\ \hline
Xception~\cite{faceforensics}   &   &\multirow{7}{*}{Image} & RGB & 73.7           & 70.9  & 72.0           & 84.5 & -              & -              \\
CNN-aug~\cite{wang2020cnn}      &                         && RGB  & 75.6           & 72.1  & 65.7           & 74.7 & -              & -              \\
Face X-ray~\cite{facexray}      &                         && RGB  & 79.5           & 65.5  & 92.8           & 86.8 & 95.4           & 84.0           \\
F3Net~\cite{F3net}              &                         && Spatial Frequency  & 68.0           & 57.9  & -              & 82.3 & 69.5           & -              \\
HFF~\cite{highfreq}             &                         && Spatial Frequency   & 74.2           & -     & 86.7           & 73.8 & 91.9           & -              \\
PEL~\cite{PEL}                  &                        && Spatial Frequency    & 69.2           & 63.3  & -              & -    & 75.9           & -              \\
SFDG~\cite{SFDG}                &                         && Spatial Frequency & 75.8           & 73.6  & -              & 92.1 & 88.0           & -              \\ 
CADDM*~\cite{caddm}                &                         && RGB & 77.6           &   73.5    &     73.8        &  87.6 &  91.3      &  80.8             \\ 
\hline
CNN-GRU~\cite{cnngru}           &    &\multirow{10}{*}{Video} & RGB   & 69.8           & 68.9  & 80.8           & 74.1 & -              & -              \\
Two-branch~\cite{twobranchrnn}  &             & & Stacked Spatial Frequency & 76.7           & -     & -              & -    & -              & -              \\
LipForensics~\cite{lips}        &  \ding{51} &        & RGB & 82.5           & 73.5  & 97.1           & 97.6 & -              & -              \\
FTCN~\cite{ftcn}               &              & & RGB  & 86.9      & 74.0  & 98.8      & 98.8 & 94.4           & 90.6           \\
MGL~\cite{MGL} &                     && Stacked Spatial Frequency           & 88.8          & -  & -  & - &  90.4         & -          \\
FCAN~\cite{NIR} &                     && Stacked Spatial Frequency & 83.5       & -     & -              & -    & -              & -          \\
RealForensics~\cite{realforensics} & \ding{51}                    & & RGB  & 86.9          & \textbf{75.9}  & \textbf{99.7}  & \underline{99.3} & 82.2*         & 88.8          \\
AltFreezing~\cite{altfreezing} &                        & & RGB & \underline{89.0} & 74.7*  & 99.2           & 99.0 & 93.7           & \underline{91.1} \\
ISTVT~\cite{ISTVT}              &                         & & RGB  & 84.2           & 74.2  & \underline{99.3} & 98.6 & -              & -    \\
StyleFlow~\cite{StyleFlow}            & \ding{51}                        & & RGB  & \underline{89.0} & 70.8* & 99.0           & 99.0 & \underline{96.1} & 90.8 \\ \hline
Ours                              &   &Video & Pixel-wise Temporal Frequency            & \textbf{89.7} & \underline{75.2} & \underline{99.3} & \textbf{99.4}  & \textbf{97.3} & \textbf{92.2} \\ \hline
\end{tabular}
}
\vspace{-2mm}
\caption{\textbf{Generalization for cross-dataset}. 
We present video-level AUC (\%) for CDF, DFDC, FSh, DFo, and DFD, with the models being trained on FF++. The highest scores are highlighted in \textbf{bold}, while secondary results are \underline{underlined}. An asterisk (*) indicates our reproduced results. The results for other methods were obtained from~\cite{StyleFlow, realforensics}.} 
  \label{tab:crossdataset}\vspace{-3mm}
\end{table*}
\begin{table*}[t]
\centering\resizebox{\textwidth}{!}{
\begin{tabular}{l|cccccc|c}
\hline
Method & HFF~\cite{highfreq} & CADDM~\cite{caddm} & FTCN~\cite{ftcn} & RealForensics~\cite{realforensics} & AltFreezing~\cite{altfreezing} & StyleFlow~\cite{StyleFlow}&\qquad Ours \qquad\qquad\\\hline
KoDF  & 85.5 & 82.7 & 88.4    & 90.4             & 90.5     &  \underline{90.7} & \qquad\textbf{91.3}\qquad\qquad \\\hline
\end{tabular}}\vspace{-2mm}
\caption{\textbf{Comparison with recent methods under racial-bias conditions.} We report AUC(\%) results evaluated on the KoDF dataset, notable for its large scale, high quality, and significant racial bias and the model trained on FF++. All results in the table are our reproduced findings.}\vspace{-5mm}
\label{fig:kodftable}
\end{table*}
\begin{table}
\centering\resizebox{0.9\columnwidth}{!}{
\begin{tabular}{l|ccc||ccc}
\hline
 \multirow{2}{*}{Method}            & Train                     & \multicolumn{2}{c||}{Test}                    & Train                    & \multicolumn{2}{c}{Test}                    \\
 & \multicolumn{1}{c}{NT}   & DF                   & FS                    & \multicolumn{1}{c}{DF}  & F2F                  & NT                   \\ \hline
\multicolumn{1}{l|}{HFF}       & \multicolumn{1}{c|}{97.9} & 88.5                 & 42.8& \multicolumn{1}{c|}{\textbf{100}} & 62.3& \underline{71.6}                   \\
\multicolumn{1}{l|}{FTCN}       & \multicolumn{1}{c|}{98.3} & \underline{97.4}                 & 90.4                & \multicolumn{1}{c|}{\underline{99.9}} & 59.2       & 66.3 \\        
\multicolumn{1}{l|}{CADDM}       & \multicolumn{1}{c|}{\textbf{99.9}} & 97.2                 & 77.6 & \multicolumn{1}{c|}{\textbf{100}} & 46.4 & 26.4                   \\      
\multicolumn{1}{l|}{StyleFlow}       & \multicolumn{1}{c|}{95.1} &            96.7    &  \underline{91.2} & \multicolumn{1}{c|}{98.7} & \underline{63.0} & 69.2                 \\  
\hline
\multicolumn{1}{l|}{Ours}       & \multicolumn{1}{c|}{\underline{99.2}}     & \multicolumn{1}{c}{\textbf{98.4}} & \multicolumn{1}{c||}{\textbf{93.4}} & \multicolumn{1}{c|}{\textbf{100}}&  \textbf{75.0}&  \textbf{82.7}\\ \hline
\end{tabular}}
\vspace{-2mm}
\caption{
\textbf{Comparison on cross-deepfake types}. We present video-level AUC (\%) performance for each deepfake type after training on the other type in FF++. All results are our reproduced findings.
}
    \label{tab:crosstype}
    \vspace{-5mm}
\end{table}
\section{Experiments}\label{sec:experiments}
We use the following forgery video datasets:
\textbf{FaceForensics++} (FF++)~\cite{faceforensics} consists of four face forgery methods (Deepfakes (DF), FaceSwap (FS), Face2Face (F2F), and NeuralTexutres (NT)).
\textbf{Celeb-DF-v2} (CDF)~\cite{CDF}, \textbf{DFDC-V2} (DFDC)~\cite{dfdc}, \textbf{FaceShifter} (FSh)~\cite{faceshifter}, \textbf{DeeperForenscis-v1} (DFo)~\cite{dfo}, \textbf{DeepFake Detection} (DFD)~\cite{dfd}, and \textbf{Korean DeepFake Detection Dataset} (KoDF)~\cite{Kwon_2021_ICCV}.
To follow the commonly used evaluation protocol of \cite{lips}, we evaluate the performance of the methods using the Area Under the receiver operating characteristic Curve (AUC) and the Equal Error Rate (EER).

We use the discrete Fourier transform with the efficient fast Fourier transform to extract temporal frequency.
We take the pre-trained TTE and 3D ResNet-50 proposed by \cite{ftcn} as 3D CNN and take the ResNet50~\cite{resnet} pre-trained for ImageNet~\cite{ImageNet} as 2D ResNet.
For training, we set a batch size of $8$ and an SGD optimizer with momentum $0.9$, $1e^{-4}$ weight decay, and a learning rate of $1e^{-4}$ for both the frequency feature extractor and STE, and a learning rate is set to $5e^{-5}$ without momentum for TTE and final classifier, and freeze the 3D CNN in the feature blender.
We train a model for 40 epochs, setting $\lambda$ to 0 for the first 4 epochs and then adjusting it to 1 for the remaining epochs.
Detailed information on datasets and implementation details will be presented in the supplementary material.

To demonstrate the effectiveness of our method, we compare it with various types of deepfake detectors, including image-based and video-based detectors. The image-based detectors are categorized into RGB-based and spatial frequency-based approaches, while the video-based detectors are categorized into RGB-based and stacked spatial frequency-based approaches. Additionally, we specified the methods utilizing extra datasets for representation learning, such as LipForensics~\cite{lips}.
Detailed information on the methods using extra datasets will be provided in the supplementary document.

\subsection{Generalization for unseen datasets}

To evaluate the generalization performance for the unseen domains, we assess a model trained on the FF++, which includes fake videos created through four different synthesis methods and original videos, across various datasets such as CDF, DFDC-v2, FSh, DFo, and DFD.
As shown in Table~\ref{tab:crossdataset}, the results indicate that even though trained only in FF++, our method demonstrates strong performance ($92.2\%$) in multiple datasets and evaluation metrics, showing exceptional generality in deepfake video detection.
Especially, the effectiveness of our pixel-wise temporal frequency is validated through experiments demonstrating that our method outperforms the methods relying on spatial frequency (e.g. FCAN, MGL). 

Additionally, to evaluate robustness under challenging conditions with significant racial bias and high quality, we conducted experiments on the KoDF dataset~\cite{Kwon_2021_ICCV}, a dataset not commonly used in previous studies. We compared our approach against several state-of-the-art methods, including HFF~\cite{highfreq}, CADDM~\cite{caddm}, FTCN~\cite{ftcn}, RealForensics~\cite{realforensics}, AltFreezing~\cite{altfreezing}, and StyleFlow~\cite{StyleFlow}.
Even with the unseen dataset of racial difference, as shown in Table~\ref{fig:kodftable}, we confirmed that our method outperformed other approaches.

\begin{figure*}[t]
  \centering
  \includegraphics[width=1\textwidth]{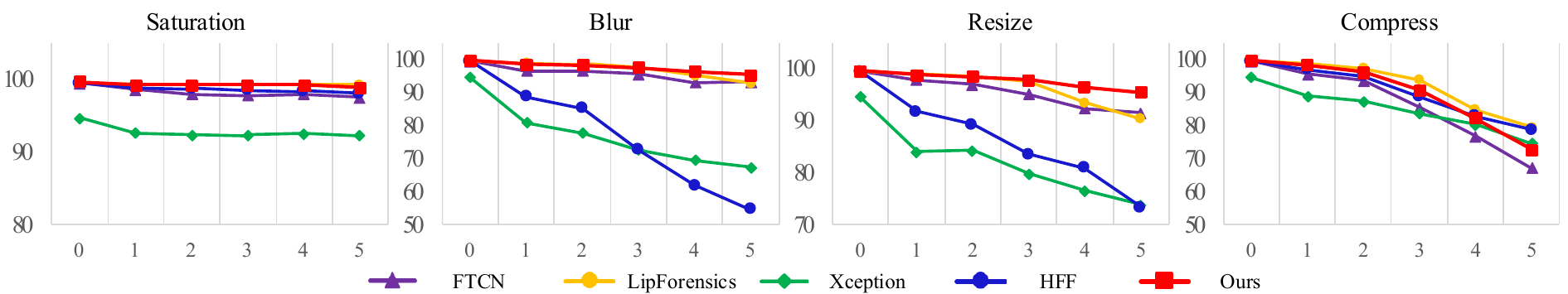} \vspace{-6mm}
  \caption{\textbf{Evaluation of the robustness against perturbation.} We compare performance across five degradation levels for four perturbation scenarios. The vertical axis shows video-level AUC (\%), and the horizontal axis represents perturbation intensity, with higher values indicating stronger perturbation. Level 0 denotes raw videos. The results in the figure are our reproductions.}
  \label{fig:perturbation} \vspace{-4mm}
\end{figure*}

\begin{table}

    \centering\resizebox{0.8\columnwidth}{!}{
    \begin{tabular}{l|cccc}
\hline
\multirow{2}{*}{Method} & \multicolumn{4}{l}{Train on remaining three}          \\ 
                        & DF   & FS   & F2F  & NT \\ \hline
    Xception ~\cite{faceforensics}               & 93.9 & 51.2 & 86.8 & 79.7  \\
    CNN-aug~\cite{wang2020cnn}                 & 87.5 & 56.3 & 80.1 & 67.8 \\
    PatchForensics~\cite{patchforensics}  & 94.0      & 60.5      & 87.3      & 84.8      \\
    HFF* ~\cite{highfreq}                    & 95.0 & 40.6 & 77.1 & 71.5 \\
    Face X-ray~\cite{facexray}            & 99.5      & 93.2      & 94.5      & 92.5   \\ 
    CNN-GRU~\cite{cnngru}                 & 97.6      & 47.6      & 85.8      & 86.6    \\
    FTCN*~\cite{ftcn}                    & \underline{99.8}      & 99.3      & 96.0      & 95.4 \\
    AltFreezing~\cite{altfreezing}   & \underline{99.8}     & \underline{99.7}      & \textbf{98.6}      & \underline{96.2}  \\ 
    \hline
    Ours                            & \textbf{99.9 }     & \textbf{99.8 }    & \underline{97.1}& \textbf{96.9}   \\\hline
    \end{tabular}}\vspace{-2mm}
        \caption{\textbf{Generalization to cross-synthesis without extra dataset}. The performances for each FF++ synthesis method, such as DF, FS, F2F and NT, after training only on the remaining three in FF++. An asterisk (*) indicates our reproduced results.}
      \label{tab:manipulation}
    \vspace{-5mm}
\end{table}

\subsection{Generalization for unseen synthesis methods}
To validate our method for unseen deepfake types, we compared it against cross-type deepfake domains (Swap $\leftrightarrow$ Reenactment) in FF++.
In Table~\ref{tab:crosstype}, when trained only on Reenactment type deepfakes (NT) and evaluated on Swap types (DF, FS), our method outperforms other methods, and vice versa (DF $\rightarrow$ F2F, NT). 
These results demonstrated that our method performs well on unseen deepfake types.

In addition to evaluating the robustness of our method for unseen synthesis, we conduct a performance evaluation with synthesis methods that are unseen during training.
We experiment with four different synthesis methods (DF, FS, F2F, NT) in FF++. 
When experimenting with one synthesis method, we evaluated it with the models trained with the training data generated by the other three methods. 
For a fair comparison of synthesis generalization ability, we compared our method with approaches that did not use extra datasets.
\begin{table}
    \centering\resizebox{\columnwidth}{!}{
    \begin{tabular}{l|cc|ccc}\hline
        Method &  FLOPs & Avg. Infer. time & CDF (EER) & DFDC (EER) & FSh (EER) \\ \hline
        FTCN & 136.2 G &14.33 ms & 0.2079 & 0.3179 & 0.0429 \\
        AltFreezing &229.4 G & 34.38 ms  & 0.2079 & \underline{0.3166} & \underline{0.0357}\\
        StyleFlow$^\ddagger$ &  4823.2 G & 1861.60 ms  & \textbf{0.1798} &0.3477 & 0.0500\\ \hline
        Ours & 192.9 G &  50.05 ms & \underline{0.1896} & \textbf{0.2919} & \textbf{0.0286}\\ \hline \multicolumn{6}{r}{ $^\ddagger$ includes pSp encoder.} \\
    \end{tabular}
    }\vspace{-3mm}
    \caption{\textbf{Deployment feasibility and EER performance}. We present the FLOPs, average inference time per clip, and the Equal Error Rate (EER) performance on the CDF, DFDC, and FSh.}
    \label{tab:deployment}
    \vspace{-3mm}
\end{table}
As shown in Table~\ref{tab:manipulation}, we demonstrate the generalization performance of our method by showing the strong performance for all unseen synthesis methods.
Our method shows remarkable effectiveness for DF and FS, as these manipulation methods produce pronounced temporal inconsistencies, which are effectively captured by our pixel-wise temporal frequency analysis.

\begin{figure}
  \begin{center}
  \includegraphics[width=1\columnwidth]{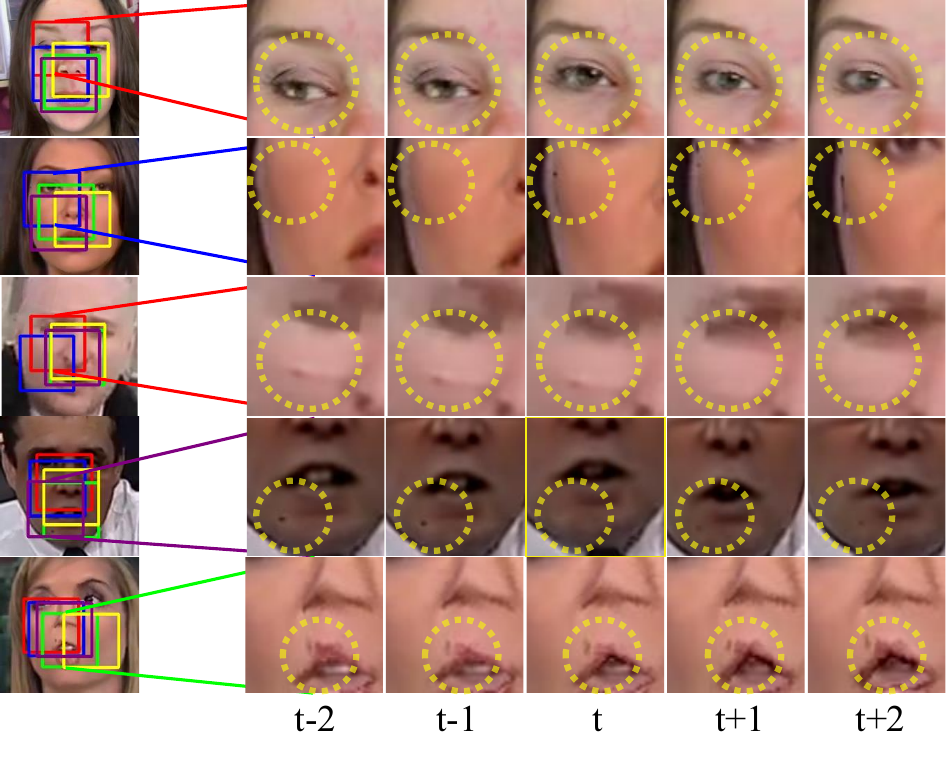}
  \end{center}
  \vspace{-9mm}
  \caption{\textbf{Visualization of part proposed by APM on extended frame.} APM propose where the temporal incoherence occurred.}
  \label{fig:facebox}
  \vspace{-6mm}
\end{figure}

\begin{figure*}[t]
  \centering
  \includegraphics[width=1\textwidth]{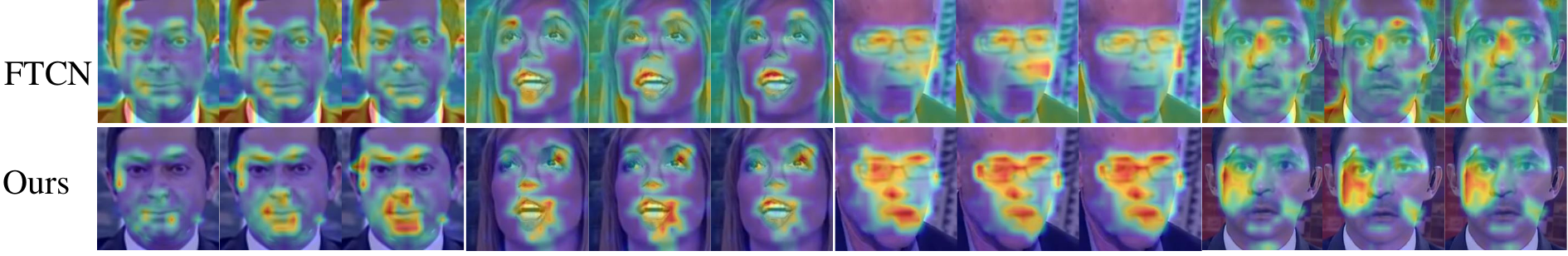}
  \vspace{-7mm}
  \caption{
  \textbf{Visualization of extended frames activation map in 3D CNN.}
  We compare activation maps from the last 3D CNN layer of FTCN (first row) and our feature blender (second row) when processing a fake clip.}
  \vspace{-5mm}
  \label{fig:activationmap}
\end{figure*}

\subsection{Robustness to Perturbations}
To evaluate robustness against perturbations, we applied saturation, Gaussian blur, resizing, and compression as outlined in \cite{dfo}. 
In Fig.~\ref{fig:perturbation}, our method shows robustness against saturation, Gaussian blur, and resizing, whereas other methods degrade in performance.
Spatial methods (HFF, Xception) suffer significant drops with blur and resizing, while temporal methods (LipForensics, FTCN) show smaller declines. Our method, which utilizes temporal frequency, sustains robustly.
On the other hand, our method maintained good performance with weak compression, but there was a limitation that the large drop as compression approached level 5.
This decline happens due to severe video compression distorting temporal information, yet our approach is still more robust than FTCN.

\subsection{Deployment Feasibility}
Our method achieves the best performance on unseen datasets while maintaining comparable computational complexity to SOTA methods, such as FTCN, AltFreezing, and StyleFlow. In Tab.~\ref{tab:deployment}, we report FLOPs and average inference time measured on an A100 GPU, showing that our approach delivers superior efficiency without incurring additional computational cost.
\subsection{Experimental Analysis}
In this section, we conduct a series of analyses to study the contribution of each component in our method. 
To analyze our method, all models are trained on FF++ and tested across FF++, DFDC, DFo, and CDF.

\subsubsection{Analysis for Frequency Feature Extractor.}

We visualize the part box determined by APM in Fig.~\ref{fig:facebox}.
The result shows that the model particularly focuses on the nose and mouth, indicating that temporal inconsistencies are prevalent in these areas.
When the APM proposed region box is used to analyze the video frame and its visualization, it is possible to detect temporal inconsistencies in the extracted region. These inconsistencies include skin tones with borders that appear and then vanish. 
We observed that the APM sometimes targets regions where temporal inconsistencies are not immediately perceptible to the human eye. 
We suggest this happens because the APM can detect subtle artifacts in the frequency domain, which may not be visually apparent in the RGB domain.

Our ablation study in Table~\ref{tab:ffe_ablation} demonstrates the crucial role of both global and part-based frequency features, as well as the effectiveness of the APM.
``Global" represents a variant using only the entire image, while ``Part'' uses only the part regions extracted by APM.
``Landmark" refers to a variant extracting five facial regions — left and right eyes, nose, and left and right mouth corners — using facial landmarks detected by RetinaFace~\cite{retinaface2020}, instead of our APM.
When either global or part-based frequency features are excluded, performance drops.
We observe that landmark-based part extraction does not contribute to improving generalization.
These findings show that global frequency features effectively detect temporal inconsistencies, while part-based frequency features enhance generalized detection by focusing on specific parts with APM.

\subsubsection{Analysis for Joint Transformer Module.}
\begin{table}
\centering
\centering\resizebox{\columnwidth}{!}{
  \begin{tabular}{lcc|ccc}
    \hline
    \multicolumn{3}{c|}{Frequency Feature Extractor} &\multicolumn{3}{c}{Testing Set} \\ 
     \multicolumn{1}{c}{Part Proposal} &  Global &\multicolumn{1}{c|}{Part}   & CDF  & DFDC  & DFo    \\ \hline     
      \multicolumn{1}{l}{Global (w/o. APM)}    &      \multicolumn{1}{|c}{\usym{2713}}  &  \multicolumn{1}{c|}{-}   & \underline{88.2}    & \underline{74.7}   &  98.9   \\
      \multicolumn{1}{l}{Part (w. APM)}    &       \multicolumn{1}{|c}{-}  & \multicolumn{1}{c|}{\usym{2713}} &   85.6 & 74.0 & 99.1   \\
      Landmark (w/o. APM)   & \multicolumn{1}{|c}{\usym{2713}}  & \multicolumn{1}{c|}{\usym{2713}}  & 81.6    & 73.8   &  \textbf{99.4}    \\
      \hline
      \multicolumn{1}{l}{Ours (w. APM)} &    \multicolumn{1}{|c}{\usym{2713}} & \multicolumn{1}{c|}{\usym{2713}} &   \textbf{89.7} & \textbf{75.2}  & \textbf{99.4} \\ \hline
    \end{tabular}}\vspace{-2mm}
    \caption{\textbf{Ablation study of Feature Frequency Extractor.} We present a comparison of different feature extraction variants: Global-based, Part-based, Landmark-based, and our integrated approach combining global and part-based frequency features extracted by APM.}
 \label{tab:ffe_ablation}\vspace{-2mm}
\end{table}

\begin{table}
\centering
\centering\resizebox{\columnwidth}{!}{
\begin{tabular}{c |c |c c |c c c}
\hline
\textbf{Architecture} & \textbf{Feature Blender} & \textbf{STE} & \textbf{TTE} & \textbf{CDF} & \textbf{DFDC} & \textbf{DFo} \\ \hline
MLP & - & - & - & 59.7 & 59.7 & 61.2 \\
MLP & ✓ & - & - & 66.6 & 69.6 & 97.6 \\
STE only & ✓ & ✓ & - & 70.7 & 71.2 & 98.0 \\
TTE only & ✓ & - & ✓ & 88.2 & 72.3 & 99.3 \\
STE + TTE (Ours) & ✓ & ✓ & ✓ & \textbf{89.7} & \textbf{75.2} & \textbf{99.4} \\ \hline
\end{tabular}%
}
    \vspace{-2mm}
    \caption{\textbf{Ablation study of Joint Transformer Module.} We present the comparison based on the inclusion of various components: the feature blender, and each transformer encoder.}
    \vspace{-5mm}
 \label{tab:jtm_ablation}
\end{table}

To see the effect of the joint transformer module, we visualize the activation map of the last layer of the feature blender compared to FTCN in Fig.~\ref{fig:activationmap}.
While FTCN targets the eyes and nose for temporal inconsistency detection, it also activates unrelated areas like clothes and background.
In contrast, our feature blender focuses solely on regions where forgery artifacts occur, like the boundaries and eyes, which explains the superior performance of our method.

Table~\ref{tab:jtm_ablation} shows the component-wise ablation studies with feature blender and varying STE and TTE in the joint transformer module.
The checkmarks indicate the presence of each module, and for the model without STE and TTE, linear classifiers were used instead. The absence of feature blender indicates that only the frequency feature extractor was used.
Even with the conventional MLP layer, the integration of frequency and RGB features through the feature blender enhances the model's generalization.
To check the effectiveness of STE and TTE, we compare the performance of classifiers trained with only the embedding values of each module.
For STE only, we trained the model for 60 epochs to get better performance.
We found that the TTE-only model shows enhanced performance on CDF compared to the STE-only model, which happens due to enhanced capturing of temporal relationships.
Finally, the combination of STE and TTE in the joint transformer shows the best performance, which is due to the effective capturing of complex relationships in each spatial and temporal dimension.
\begin{table}
  \centering
  \begin{tabular}{l|ccc|c}
\hline
Training Strategy                               & CDF & DFDC   & DFo & Avg.\\ \hline
Ours & \textbf{89.7}  & \textbf{75.2}   & \textbf{99.4} & \textbf{88.1}\\ 
{\hspace{5mm}\small $\triangleright$ No $\mathcal{L}_{\text{auxiliary}}$} &  \underline{87.2}   &  \underline{74.8}  & \textbf{99.4} & \underline{87.1}  \\ 
{\hspace{5mm}\small $\triangleright$ $\lambda = 1$ at every epoch}                & 86.9  & 74.6  & \underline{99.3}  &   86.9  \\
\hline
\end{tabular}
    \vspace{-2mm}
\caption{\textbf{Ablation study for training phase.} We present an analysis of the training phase. This table presents the impact of auxiliary loss and the setting of $\lambda = 1$ at every epoch.
  }
  \label{tab:phase_ablation}
  \vspace{-5mm}
\end{table}
\subsection{Analysis for Training Phase.}
We analyze the training phase in Table~\ref{tab:phase_ablation}. 
We observe a performance drop when the auxiliary loss is not considered. The auxiliary loss helps train the APM and spatial transformer encoder more effectively.
When $\lambda = 1$ is initially set (as shown in Eq.~\ref{equ:phase2}), we observe comparable performance; however, setting $\lambda = 1$ after the first 4 epochs allows the weights in APM and STE to align and be trained in an integrated form.
\section{Conclusion}
\label{sec:conclusion}

In this paper, we present a novel forgery detection approach based on pixel-wise temporal frequency.
We first demonstrate that temporal frequency can be used to detect forgery and then use experiments to show how it can help where other methods fail.
Contrary to spatial frequency, pixel-wise temporal frequency can detect local temporal inconsistency, which makes generalized deepfake video detection possible.
We also propose a framework to fuse temporal frequency information with RGB video information.
Finally, we perform forgery detection through the automatic mechanism of extracting the region of interest and our solution is more robust and generalized than previous methods. 

\vspace{1mm}
\footnotesize
\noindent\textbf{Acknowledgements:}
This work was partly supported by Institute of Information \& Communication Technology Planning \& Evaluation (IITP) grant funded by the Korea government (MSIT) (No. RS-2025-02263841, Development of a Real-time Multimodal Framework for Comprehensive Deepfake Detection Incorporating Common Sense Error Analysis; RS-2021-II211341, Artificial Intelligence Graduate School Program (Chung-Ang University); No. RS-2020-II201336 AIGS program (UNIST)).
\normalsize
\clearpage
\setcounter{page}{1}
\setcounter{figure}{0}
\setcounter{equation}{0}
\setcounter{table}{0}
\setcounter{section}{0}
\renewcommand{\thefigure}{\Alph{figure}}
\renewcommand{\thetable}{\Alph{table}}
\renewcommand{\theequation}{\arabic{equation}s}
\renewcommand{\thesection}{\Alph{section}}
\maketitlesupplementary

\begin{figure}
  \centering
  \includegraphics[width=.8\columnwidth]{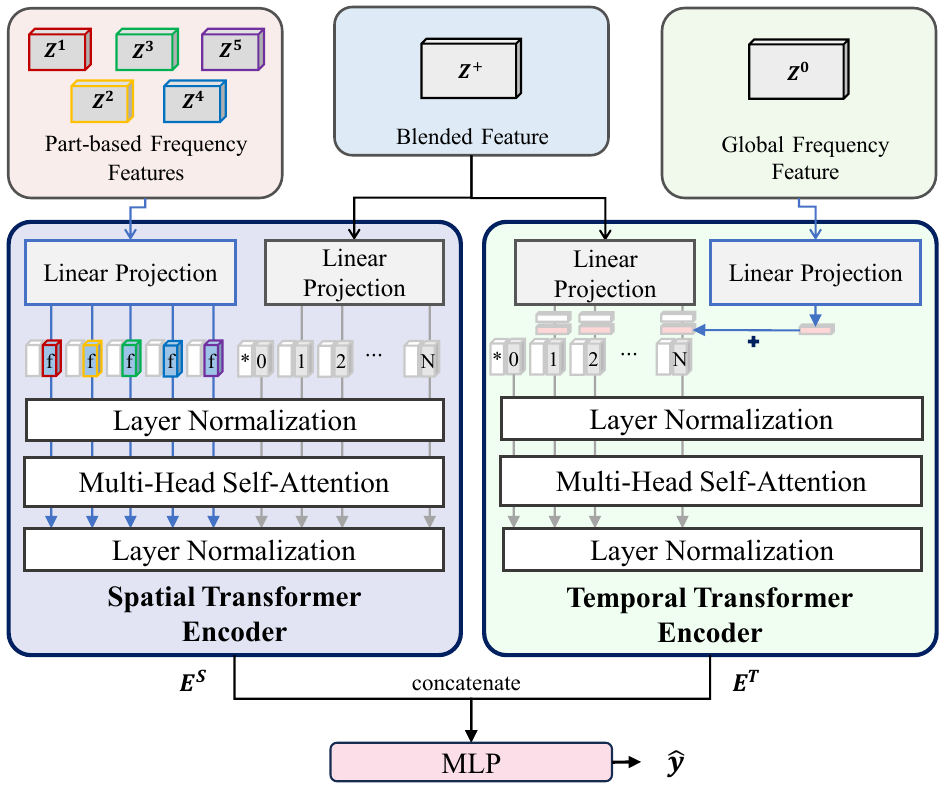}
  \caption{\textbf{Visualization of Transformer-based Integration architecture.} The transformer-based integration takes the global frequency feature $z^{0}$, the part-based frequency features $z^{p},\quad p\in\{1,2,3,4,5\}$, and the blended feature $Z^{+}$. The transformer-based integration obtains the spatial embedding $E^S$ using spatial transformer encoder and the temporal embedding $E^T$ using temporal transformer encoder and uses these embeddings to make the final prediction $\hat{y}$. For STE and TTE, the frequency features are subjected to average pooling, producing $Pool(z^x)\in R^{B\times C}$, with $x\in \{0,...,5\}$.}
  \label{fig:jointtransformer}
\end{figure}
\section{Detail of Transformer-based Integration}
In this section, we will describe in detail the structure of the transformer-based integration in the joint transformer module introduced in the previous section.
Our transformer-based integration is mainly composed of a Spatial Transformer Encoder (STE) and a Temporal Transformer Encoder (TTE), and we visualize its architecture in Fig.~\ref{fig:jointtransformer}.

\subsection{Spatial Transformer Encoder}
We design the spatial transformer encoder to improve the interaction between the spatial feature from the feature blender and the part-based frequency features.
STE outputs the spatial embedding $E^{S} \in R^{1024}$ by feeding the spatial features of the blended feature $Z^{0}\in R^{D_C\times D_T\times D_H\times D_W}$ and each part-based frequency feature $Z^{p}$ as a token to the Standard Transformer.

We first estimate the spatial features $Z^{sp}\in R^{D_C\times 1\times D_H\times D_W}$ from blended feature $Z^{+} $ by averaging on the temporal axes, and STE generates the spatial embedding $E^{S}$ by getting $Z^{sp}$ as a token.
We apply linear projection $W^{sp}$ to map the sequence of features $z^{sp}_s \in R^{D_C},s\in\{1,2,...,D_H\times D_W\}$ of $Z^{sp}$ and add a 2D positional encoding $pos^{sp}$.
\begin{equation}
         tokens^{sp}_{+} = [z^{sp}_{class}, W^{sp}z^{sp}_1, ..., W^{sp}z^{sp}_{D_H\times D_W} ]^T + pos^{sp},
    \label{equ:spatial_embedding}
\end{equation}
where $pos^{sp} $ is 2D sincos positional encoding, $Z^{sp}_{class}$ is extra class embedding, and `$sp$' is short for spatial. 
Using the coordinate values ($a_p, b_p$) employed to crop each part-based frequency feature, the $p^{th}$ part position encoding value $pos^{part}_p$ is obtained by interpolating neighboring positional encoding values in $pos^{sp}$ and then adding it to a frequency position value $pos^{freq}$. 
This frequency position value $pos^{freq}$ serves as a specific indicator of frequency domain features.
\begin{equation}
    pos^{part}_{p} =  \text{interpolation}((a_p,b_p), pos^{sp}) + pos^{freq} \small, p\in \{1,...,5\}.
\end{equation}

We apply linear projection $W^{freq}$ to map the sequence of part-based frequency features $Z^{p}$ and add with part position $pos^{part}_{p}$. 
\begin{equation}
    tokens^{sp}_{freq}  = [W^{freq}Z^{1} + pos^{part}_{1},  \ldots,  W^{freq}Z^{5} + pos^{part}_{5}]^T.
\label{eq:freq_embedding}
\end{equation}

We concatenate the tokens $tokens^{sp}_{+}$ and $token^{sp}_{freq}$, which are fed into a transformer to get the spatial transformer embedding $E^{S}$.
\begin{equation}
    E^{S} = \mathbf{STE}(tokens^{sp}_{+}, tokens^{sp}_{freq}).
    \label{STE}
\end{equation}

\subsection{Temporal Transformer Encoder}
Similar to STE, the Temporal Transformer Encoder (TTE) takes the temporal features $Z^{tp}\in R^{D_T\times D_C\times 1\times 1}$ of the blended feature $Z^{+}$ and the global frequency features $Z^{0}$ and outputs the temporal embedding $E^{T} \in R^{1024}$.
To make token $tokens_{tp}$, we apply linear projection $W^{tp}$ to $Z^{tp}_t \in R^{D_C},t\in\{1,2,...,D_T\}$ and add with 1D positional encoding $pos^{tp}$.
$Z^{tp}_{class}$ is extra class embedding and `tp' is short for temporal. 
\begin{equation}
    tokens^{tp} = [Z^{tp}_{class}, W^{tp}Z^{tp}_1, ..., W^{tp}Z^{tp}_{D_T} ]^T\nonumber + pos^{tp}.
    \label{equ:token_tp}
\end{equation}

We put $tokens^{tp}$ into a transformer to get the temporal transformer embedding $E^{T}$ after adding with mapped global frequency feature by linear projection $W^{tp}_{freq}$.
\begin{equation}
    E^{T} = \mathbf{TTE}(tokens^{tp}+ W^{tp}_{freq} Z^{0} ).
    \label{STE}
\end{equation}
To get the final prediction $\hat{y}$, we concatenate $W^{b} E^{S}$ and $E^{T}$, which are put into the final classifier $\phi^{final}$.
\begin{equation}
    \hat{y} = \phi^{final}(W^{b} E^{S},E^{T}),
    \label{mlpfinal}
\end{equation}
where $W^{b} \in R^{1024}$ is a linear projection that aligns distributions.

\section{Detail of Updating Attention Proposal Module.}
\begin{figure}[t]
  \begin{center}
  \includegraphics[width=0.8\columnwidth]{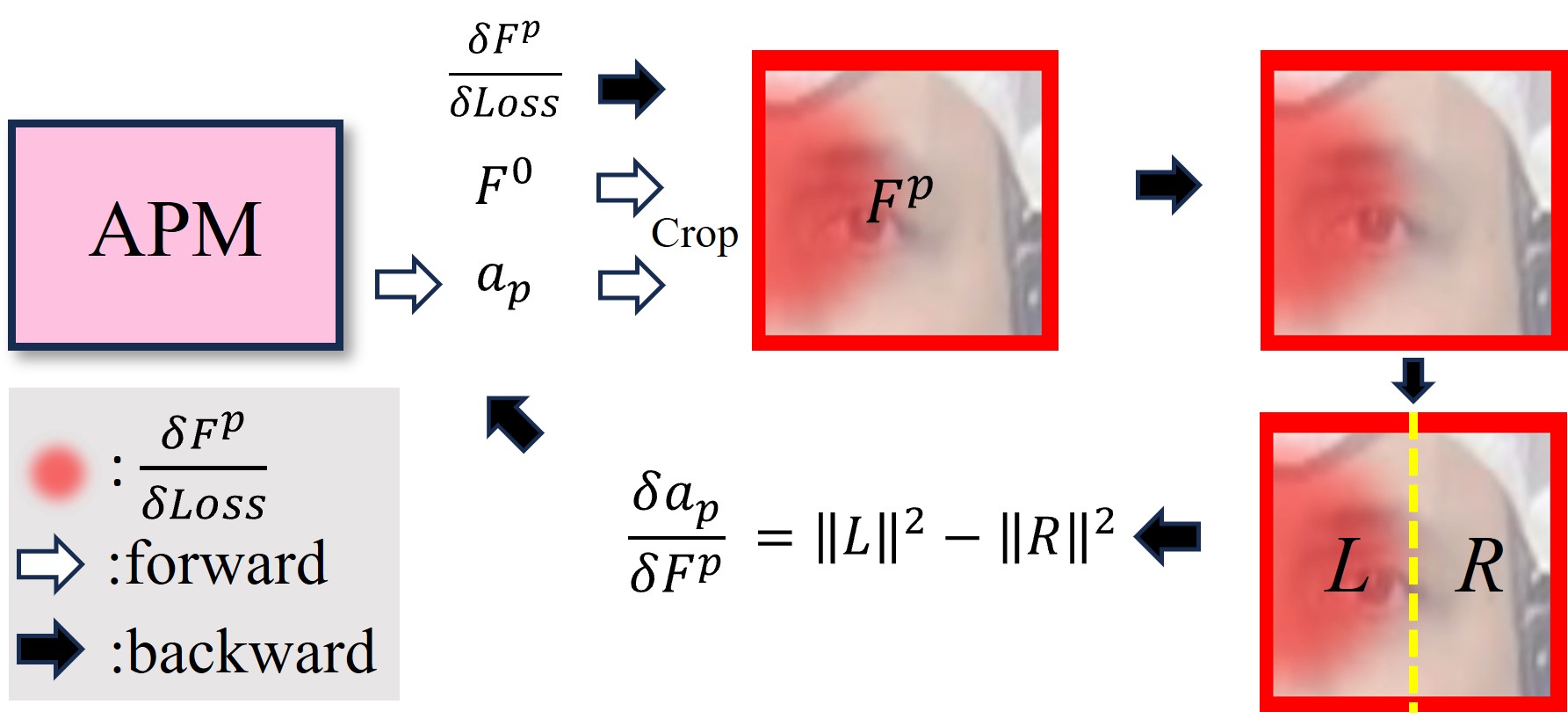}
  \end{center}
  \caption{\textbf{Visualization of Updating the Attention Proposal Module.} We visualize the process of updating APM.}
  \label{fig:apmupdate}
\end{figure}

This section details the operation of the Attention Proposal Module (APM) within the Frequency Feature Extractor (introduced in Section 4.1). 
The APM automatically identifies and focuses on artifact-rich regions within the input, enabling adaptive localization of deepfake manipulations. Unlike methods relying on predefined regions (e.g., facial landmarks), the APM leverages classification gradients to pinpoint areas most relevant for deepfake detection.

To update the parameters of the APM, the following steps are performed. First, the input frequency spectra are masked by applying $M_p(a_p,b_p)$ according to Eq. 4, ensuring proper gradient computation during backpropagation. The regions corresponding to elements with a value of 1 in $M_p$ are then cropped for part-based frequency extraction. Importantly, this cropping operation does not interfere with gradient propagation, as masked-out regions (with zero values) naturally propagate zero gradients.
Next, gradients derived from the binary classification loss are backpropagated through the cropped regions, guided by the APM's outputs $a_p$ and $b_p$ in Eq. 3.
Taking the x-axis as an example, the gradients are divided into left $L$ and right $R$ segments ($R, L \in \mathbb{R}^{H \times \frac{W}{2}}$). The squared magnitudes of the gradient values ($||L||^2$ and $||R||^2$) corresponding to each segment are calculated and their difference $||L||^2 - ||R||^2$ is computed and transmitted as the gradient value for $a_p$. Thus, if the gradient magnitude is larger on the left segment, the APM is encouraged to propose a smaller x value (to the left). Similarly, $b_p$ is updated along the corresponding axis. Through this training procedure, the APM learns to effectively focus on regions exhibiting prominent gradient responses (artifact-rich areas), thereby enhancing its ability to localize and analyze deepfake anomalies.
We describe this process in Fig.~\ref{fig:apmupdate}.

\section{Analysis of Temporal Frequency Extraction}

\begin{figure}
  \centering
  \includegraphics[width=1\columnwidth]{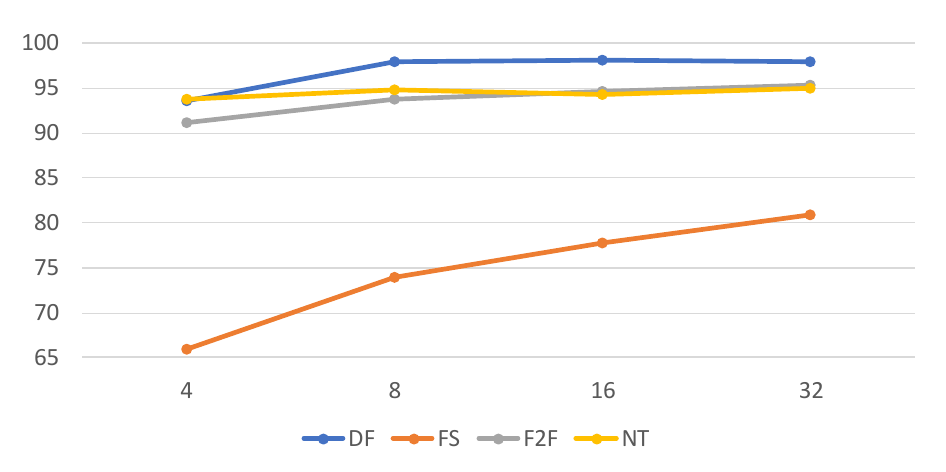}
  \caption{\textbf{Comparison of frame intervals for temporal-frequency extraction.} The horizontal axis is the frame interval used to extract the temporal frequency and the vertical axis is the video-level AUC(\%) performance of the cross-synthesis experiments for each method.}
  \label{fig:freqstride}
\end{figure}

We experimented to check how to extract temporal frequency for deepfake video detection. 
Table~\ref{tab:freqAnalysis} and Fig.~\ref{fig:freqstride} are the results of a cross-synthesis experiment using the ResNet-50~\cite{resnet} classifier trained by global temporal frequency only.
\begin{table}[h]
    \centering
    \subfloat[\textbf{Filtering Method.}]{
        \begin{tabular}{|l|ccccc|}
            \hline
            \multirow{2}{*}{Filter} & \multicolumn{5}{c|}{Train on remaining three}          \\ \cline{2-6} 
                                     & DF   & FS   & F2F  & \multicolumn{1}{c|}{NT}   & Avg  \\ \hline
            None                     & 73.01 & 53.26 & 64.40 & \multicolumn{1}{l|}{58.33} & 62.25 \\ 
            Median                   & \textbf{98.05} & \textbf{80.98} & \textbf{95.29} & \multicolumn{1}{l|}{\textbf{95.02}} & \textbf{92.34} \\ 
            Mean                     & 96.85 & 78.13 & 90.88 & \multicolumn{1}{l|}{93.39} & 89.81 \\ 
            \hline
        \end{tabular}
        \label{tab:freqA}
    }
    
    \subfloat[\textbf{Comparison of Frequency Features.} ]{
        \begin{tabular}{|l|ccccc|}
            \hline
            \multirow{2}{*}{Feature} & \multicolumn{5}{c|}{Train on remaining three}          \\ \cline{2-6} 
                                     & DF   & FS   & F2F  & \multicolumn{1}{c|}{NT}   & Avg  \\ \hline
            Magnitude                & \textbf{98.05} & \textbf{80.98} & \textbf{95.29} & \multicolumn{1}{l|}{95.02} & \textbf{92.34} \\ 
            Phase                    & 54.36 & 50.82 & 55.25 & \multicolumn{1}{l|}{60.93} & 55.34 \\ 
            All                      & 95.45 & 73.22 & 92.81 & \multicolumn{1}{l|}{\textbf{95.96}} & 89.36 \\ 
            \hline
        \end{tabular}
        \label{tab:freqB}
    }
    \caption{\textbf{Experiment about temporal frequency extraction methods.}} 
    \label{tab:freqAnalysis}
\end{table}

For the experiment using one synthesis method, we evaluated our method by training a model with the training data generated by the other three methods. We experiment with four different synthesis methods (DF, FS, F2F, NT) in FaceForensics++~(FF++)~\cite{faceforensics}.

In our default setting, as a pre-processing, we apply the median filter to extract the frequency for 32 frames, using the magnitude of the filtered frequency to detect the deepfake video.
Before performing 1D Fourier transform, we obtain a pre-processed frame $\hat{I}$ by removing the dominant components through a median filter as 
\begin{equation}
    \hat{I} = gray(I - filter(I)),
    \label{equ:sup_filter}
\end{equation}
where $filter(I)$ is an image filtered by a median filter and $gray(I)$ means the gray-scaling function from the colored image $I$.

We compare the performance of the process with and without a filter and the performance of different types of filters and show the results in Table~\ref{tab:freqA}.
The first row (None) is the result without filter, which means the result when the original image $I$ is gray-scaled and Fourier transformed and fed into ResNet-50, the second row (Median) is the result when median filter is applied as filter, and the last row (Mean) is the result when Mean filter is applied in Eq.~\ref{equ:sup_filter}.
Even though the mean filter was also effective for our method, the median filter showed the best performance.

Table~\ref{tab:freqB} is the result when using phase and magnitude of temporal frequency.
Table~\ref{tab:freqB} presents the performance for different frequency features—Magnitude, Phase, and a combination of both (All).
We observe that employing magnitude outperforms employing phase alone.

Fig.~\ref{fig:freqstride} is a performance table for each frame interval over which frequencies are extracted. For example, an interval of 4 means that the temporal frequency was extracted every 4 frames for a total of 32 frames.
We find that performance increased with higher frame intervals but saturated at 8 for all, except for FS.

Through these experiments, we use the Fourier transform to extract the temporal frequency magnitude from 32 frames preprocessed by a median filter.

\section{Additional Analysis}

\subsection{Clarification of the Use of Temporal Information}
To demonstrate that our pixel‐wise temporal-frequency-based approach focuses on temporal information rather than encoding static appearance, we randomly permuted the order of the 32‐frame input sequence and evaluated its resulting performance.
As shown in Tab.~\ref{tab:shuffled_performance}, our method shows a significant performance drop (i.e., -43.8\%), compared to AltFreezing, demonstrating the strong dependency on the temporal information. These results confirm that, rather than encoding static appearance cues, our approach focuses on temporal dynamics.
\begin{table}
    \centering\resizebox{\columnwidth}{!}{
    \begin{tabular}{l|c|cc}\hline
        Method & FF++  & CDF& FSh \\ \hline
        AltFreezing    & 99.7\,→\,62.2 ($-$37.6 \%) & 89.0\,→\,56.8 ($-$36.2 \%) & 99.0\,→\,63.2 ($-$36.2 \%) \\
        Ours           & 99.6\,→\,56.0 ($-$43.8 \%) & 89.7\,→\,57.8 ($-$35.6 \%) & 99.4\,→\,53.1 ($-$46.6 \%) \\
        \hline
    \end{tabular}
    }
    \caption{\textbf{Performance degrade on shuffled frames.}}
    \label{tab:shuffled_performance}
\end{table}

\subsection{Effectiveness of APM.}
\begin{table}
\centering
\resizebox{0.8\columnwidth}{!}{
  \begin{tabular}{l|c|ccc|l}
    \hline 
    Arch. &  APM  &FSh   & DFDC  & DFo & Avg.   \\ \hline
     MLP & &  85.0  & 64.0  & 88.6  & 79.2 \\  
     MLP & \usym{2713}&   97.4 & 69.6  & 97.6     & 88.2 (11.4\% $\uparrow$)\\ \hline
    \end{tabular}
    }
    \caption{\textbf{Isolated Effectiveness of the APM.} We trained the model on FF++ and evaluated it for FSh, DFDC and DFo.}
   
 \label{tab:apmablation}
\end{table}

To evaluate the effectiveness of the APM in isolation, independent of modules like STE and TTE, we conducted experiments as shown in Table~\ref{tab:apmablation}. These experiments clearly demonstrate that incorporating the APM significantly improves performance. Specifically, the model without APM resulted in a lower performance, adding the APM alone led to a performance increase of up to 11.4\%.

\begin{figure}[t]
  \centering
  \includegraphics[width=1\columnwidth]{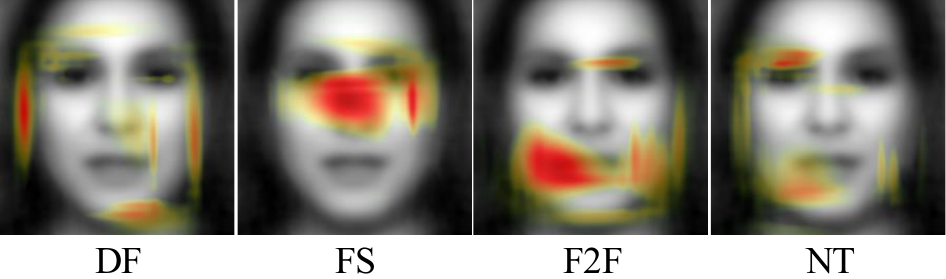}
  \caption{\textbf{APM Over-Selection Heatmaps (red: regions selected more frequently than average) for each deepfake type.}}
  \label{fig:rebuttal_apmdistribution}
\end{figure}

In Fig.~\ref{fig:rebuttal_apmdistribution}, we further analyzed APM activations by visualizing normalized heatmaps of APM regions for DF, FS, F2F, and NT. Regions are meaningfully different according to datasets, denoting no overfitting.
Specifically, APM tends to focus on facial parts that were previously helpful to reveal deepfakes: the mouth for F2F, eyes and nose for FS.

\begin{figure}[t]
  \centering
  \includegraphics[width=1\columnwidth]{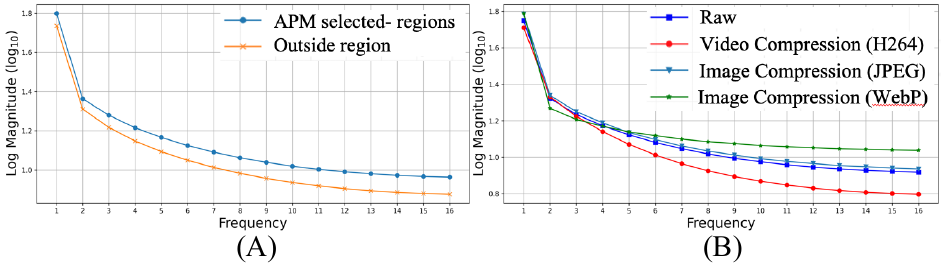}
  \caption{\textbf{Pixel‐wise Temporal Frequency Distributions in FF++}: (A) shows inside APM-selected patches versus the remaining area, (B) presents under different compression. }
  \label{fig:frequency}
\end{figure}

In Fig.~\ref{fig:frequency}-A, APM-selected patches show higher average temporal frequency magnitudes than surrounding areas, indicating stronger flickering effects.

\begin{table}[t]
    \centering\resizebox{\columnwidth}{!}{
    \begin{tabular}{l|ccccc|c}
\hline
The number of parts & CDF        & DFDC & FSh           & DFo           & DFD        & Avg.       \\ \hline
0                   & 88.2       & 74.7 & \textbf{99.4} & \underline{98.9}    & 96.8       & 91.6       \\
1                   & \underline{89.3} & 74.7 & \underline{99.3}    & \textbf{99.4} & 96.9       & \underline{91.9} \\
3        & 88.4          & \textbf{75.3} & \underline{99.3} & \textbf{99.4} & \textbf{97.3} & \underline{91.9}    \\
5 (ours) & \textbf{89.7} & \underline{75.2}    & \underline{ 99.3} & \textbf{99.4} & \textbf{97.3} & \textbf{92.2} \\
7                   & 87.1       & 74.5 & \underline{99.3}    & \textbf{99.4} & \underline{ 97.0} & 91.5       \\ \hline
\end{tabular}}
    \caption{\textbf{Performance comparison by each number of parts proposed by APM.} We trained the model on FF++ and evaluated it for five unseen datasets.}
    \label{tab:nun_parts}
\end{table}

\subsection{Ablation Study for the number of parts}
Table~\ref{tab:nun_parts} shows the performance of a model trained with varying numbers of parts extracted by APM across different datasets.
We observe performance improvements after incorporating part-based frequency features from APM, particularly with a notable increase in performance on the DFo dataset, generally increasing with up to 5 parts and peaking at 5.
However, performance declined with too many parts due to redundancy from observing similar regions repeatedly, with noticeable drops in performance on the DFDC and CDF datasets.

\subsection{Ablation Study for the Feature Blending.}
To analyze methods for integrating raw RGB features with temporal frequency features, we conducted experiments comparing different integration strategies in Table~\ref{tab:featureblender}. Our findings show that using $1\times 1$ Conv. for feature blending improves performance while omitting it reduces adaptation between RGB and temporal frequency domains, leading to a drop (e.g. no blending, add, concatenate). Additionally, increasing convolution depth offers no significant gain.

\begin{table}
\centering
\centering\resizebox{0.8\columnwidth}{!}{
  \begin{tabular}{l|cccc|c}
    \hline 
    Feature blending& FSh   & CDF  & DFDC  & DFo& Avg.    \\ \hline
    no blending& 99.3 & 85.1& 75.4 & 99.3  & 89.8 \\
    add  & 77.3 & 68.2& 59.9 & 75.7 & 70.3 \\
    concatenate  & 99.0 & 85.4 & 74.4& 99.4 &89.6 \\
    
    $1\times 1$ Conv. (Ours)  &99.3 &\textbf{89.7}  & 75.2 & 99.4& \textbf{90.9}  \\ 
      $1\times1$ Conv. ($\times 2$) &\textbf{99.4} & 87.1   &  \textbf{75.6} &\textbf{99.5}& 90.4    \\
      \hline
    \end{tabular}
    }
    \caption{\textbf{Analysis for feature blending methodologies.} We trained the model on FF++.}
 \label{tab:featureblender}
\end{table}

\begin{table*}
  \centering
  \begin{tabular}{l|cccccccc|c}
\hline
Method       & Clean & Saturation & Contrast & Block & Noise & Blur & Resize & Compress & Avg.  \\ \hline
Xception     & 99.8  & 99.3       & 98.6     & 99.7  & 53.8  & 60.2 & 74.2  & 62.1     & 78.3 \\
CNN-aug      & 99.8  & 99.3       & 99.1     & 95.2  & 54.7  & 76.5 & 91.2  & 72.5     & 84.1 \\
Patch-based  & 99.9  & 84.3       & 74.2     & 99.2  & 50.0  & 54.4 & 56.7  & 53.4     & 67.5 \\
CNN-GRU      & 99.9  & 99.0       & 98.8     & 97.9  & 47.9  & 71.5 & 86.5  & 74.5     & 82.3 \\
LipForensics* & 99.6  & 99.3       & 98.8     & 98.7  & 64.3  & 96.7 & 95.8  & 90.9     & 92.1 \\ 
FTCN*        & 99.5  & 98.0       & 93.7     & 90.1  & 53.8  & 95.0 & 94.8  & 83.7     & 87.0 \\ 
\hline Ours        & 99.7   & 99.1      & 95.8    & 91.9  &  55.0  & 97.3 & 97.5  &  88.0      &  89.2  \\ \hline
\end{tabular}
    \caption{\textbf{Average robustness for perturbations.} We present video-level AUC(\%) for each perturbation.}
  \label{tab:perturbation}
\end{table*}
\begin{figure}
  \centering
  \includegraphics[width=1\columnwidth]{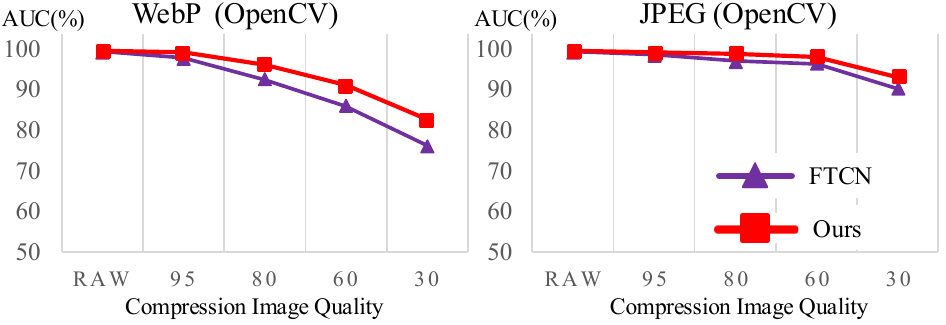}
  \caption{\textbf{Evaluation of the robustness against WebP/JPEG.}}
  \label{fig:rebuttal_pertur}
\end{figure}
\subsection{Additional Results for Perturbations Robustness.}
We conducted a study on the robustness of our model against various perturbations as proposed in DFo~\cite{dfo}.
As shown in Table~\ref{tab:perturbation}, our model shows impressive robustness against perturbations like resize and blur. This robustness indicates that our model is robust for frame-wise perturbation.
Meanwhile, our method suffers from performance drops for perturbations that distort temporal information, such as compression and noise. These performance drops are due to temporal distortion interfering with the extraction of temporal frequency.

To evaluate our approach on more general compression, we conducted additional compression robustness experiments on WebP and JPEG compression. As shown in Fig.~\ref{fig:rebuttal_pertur}, under severe degradation using WebP and JPEG compression, our method demonstrates superior robustness.

\section{Discussion.}
\begin{wrapfigure}{r}{0.4\columnwidth}  
  \centering
\vspace{-4mm}
  \includegraphics[width=0.4\columnwidth]{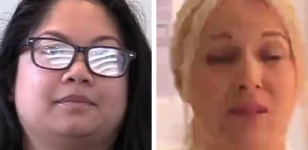}
  \vspace{-7mm}
  \caption{\textbf{Failure Case.}}
  \label{fig:failure}
\vspace{-7mm}
\end{wrapfigure}

Fig.~\ref{fig:failure} denotes a failure case in which specular reflections on eyeglass lenses lead the model to misclassify genuine frames as forgeries.
Moreover, as shown in Fig.~4 of the main paper and Fig.~\ref{fig:rebuttal_pertur}, the temporal frequency generally remains robust under various video manipulations, such as resizing and saturation adjustments. However, heavy video and image compression can merge multiple neighboring pixels into a single subpixel, diminishing pixel-level motions and amplifying domain shifts in the temporal frequency.

Fig.~\ref{fig:frequency}-B compares the average pixel-wise temporal frequency under diverse image and video compression methods (H264, JPEG, and WebP). At low frequency, compressed images/videos follow an uncompressed signal (i.e., RAW) closely; while at high frequency, compressed images/videos increasingly diminish the spectrum. This attributes the performance drop to ours under severe compression.
The performance drop under severe compression remains a limitation, and our future work may deal with this by exploring temporal-frequency regularization.

\section{More Detail Setting}
In this section, we will present the detailed setup of our experiments.

\subsection{Datasets}
We use the following forgery video datasets:
(1) \textbf{FaceForensics++} (FF++)~\cite{faceforensics} consists of four face forgery methods (Deepfakes (DF), FaceSwap (FS), Face2Face (F2F), and NeuralTexutres (NT)), with a total of 5000 videos (1000 real videos and 1000 fake videos for each method).
(2) \textbf{Celeb-DF-v2} (CDF)~\cite{CDF} is a dataset consisting of 590 real videos and 5,639 fake videos that are synthesized by advanced techniques compared to FF++.
(3) \textbf{DFDC-V2} (DFDC)~\cite{dfdc}, which has a 3,215 test video set, is more challenging than other datasets and was made under extreme conditions.
(4) \textbf{FaceShifter} (FSh)~\cite{faceshifter} and (5) \textbf{DeeperForenscis-v1} (DFo)~\cite{dfo} contains high-quality forgery videos generated from the real videos from FF++.
(6) \textbf{DeepFake Detection} (DFD)~\cite{dfd} is a popular benchmark dataset for forgery detection with 363 real videos and 3071 synthetic videos.
(7) \textbf{Korean DeepFake Detection Dataset} (KoDF)~\cite{Kwon_2021_ICCV} dataset is composed of 62,166 real videos and 175,776 fake videos generated using six face forgery methods. We employed a validation set for our experiments and utilized the initial 110 frames from each video for analysis.

\subsection{Comparisons}
To demonstrate the effectiveness of our method, we compare it with various types of deepfake detectors, including image-based and video-based detectors. 
Image-based detectors can be categorized into RGB-based and spatial frequency-based approaches. RGB-based detectors use RGB images to identify deepfakes, while spatial frequency-based approaches apply filters or Fourier transforms to extract spatial frequency, combining these with RGB information to improve detection.

Similarly, video-based detectors are categorized as RGB-based and stacked spatial frequency-based approaches. RGB-based methods in video detectors use individual RGB video frames for deepfake detection, whereas stacked spatial frequency-based approaches extract spatial frequencies (similar to the image-based frequency methods) and stack them temporally for enhanced detection.

Also, we categorize certain approaches based on whether they utilize external datasets. Specifically, LipForensics~\cite{lips} and RealForensics~\cite{realforensics} incorporate the LRW~\cite{lrw} dataset to learn representations of lip or facial motions, while StyleFlow~\cite{StyleFlow} leverages a PsP encoder~\cite{psp} pre-trained on the FFHQ~\cite{ffhq} dataset to extract style latent.

We have reproduced several methods, including FTCN~\cite{ftcn}, CADDM~\cite{caddm}, HFF~\cite{highfreq}, LipForensics~\cite{lips}, RealForensics~\cite{realforensics}, AltFreezing~\cite{altfreezing}, and StyleFlow~\cite{StyleFlow}.
When available, we used their pre-trained weights for performance comparison, obtained from the official GitHub repositories.
If the official weights did not align with our experimental settings, or if the official weights were unavailable—such as for CADDM, which was trained on FF++ with FSh—we retrained the models according to our specific experimental setup.

We report performance using Video-level AUC, which is calculated by averaging the clip-level predictions for each clip input sequence within a video to obtain a single video-level prediction. The AUC is then computed based on these aggregated video-level predictions.

\subsection{Implementation Details}
We use RetinaFace~\cite{retinaface2020} to detect faces and perform face tracking with SORT~\cite{sort}. 
When cropping faces, we picked a region based on the average point of the landmarks of the faces over 32 frames, cropped the same area for 32 frames, and fed to our model these consecutive 32 frames as input. 

Most experiments were conducted using four Nvidia A6000 48GB GPUs and an AMD Ryzen Threadripper Pro 3955WX 16-Cires CPU or two A100 40GB GPUs and an Intel Xeon Gold 6240R CPU, while experiments using only 2D ResNet, such as Table~\ref{tab:freqAnalysis}, were conducted using four Nvidia RTX-3090 24GB GPUs and Intel i9-10900X CPU.
{
    \small
    \bibliographystyle{ieeenat_fullname}
    \bibliography{main}
}
\end{document}